 \documentclass[smallabstract,smallcaptions]{dccpaper}

\usepackage{epsfig}
\usepackage{amsmath}
\usepackage{amssymb}
\usepackage{color}
\usepackage{url}
\usepackage{subcaption}
\usepackage{multirow}
\usepackage{wrapfig}
\usepackage{graphicx}
\newlength{\figurewidth}
\newlength{\smallfigurewidth}

\begin{document}

\title
{\large
\textbf{Lightweight 3D Gaussian Splatting Compression \\via Video Codec}
}

\author{%
Qi Yang$^{\ast}$, Geert Van Der Auwera$^{\dag}$, and Zhu Li$^{\ast}$\\[0.5em]
{\small\begin{minipage}{\linewidth}\begin{center}
\begin{tabular}{ccc}
$^{\ast}$ University of Missouri - Kansas City & \hspace*{0.5in} & $^{\dag}$Qualcomm \\
\end{tabular}
\end{center}\end{minipage}}
\thanks{
This work is supported in part by an award from NSF 2148382, and a gift grant from Qualcomm. Corresponding Author: Qi Yang ({qiyang@umkc.edu})
}
}
\maketitle
\thispagestyle{empty}

\begin{abstract}
 Current video-based GS compression methods rely on using Parallel Linear Assignment Sorting (PLAS) to convert 3D GS into smooth 2D maps, which are computationally expensive and time-consuming, limiting the application of GS on lightweight devices. In this paper, we propose a \textbf{L}ightweight 3D Gaussian Splatting (GS) \textbf{C}ompression method based on \textbf{V}ideo codec (LGSCV). First, a two-stage Morton scan is proposed to generate blockwise 2D maps that are friendly for canonical video codecs in which the coding units (CU) are square blocks. A 3D Morton scan is used to permute GS primitives, followed by a 2D Morton scan to map the ordered GS primitives to 2D maps in a blockwise style. However, although the blockwise 2D maps report close performance to the PLAS map in high-bitrate regions, they show a quality collapse at medium-to-low bitrates. Therefore, a principal component analysis (PCA) is used to reduce the dimensionality of spherical harmonics (SH), and a MiniPLAS, which is flexible and fast, is designed to permute the primitives within certain block sizes. Incorporating SH PCA and MiniPLAS leads to a significant gain in rate–distortion (RD) performance, especially at medium and low bitrates. MiniPLAS can also guide the setting of the codec CU size configuration and significantly reduce encoding time. Experimental results on the MPEG dataset demonstrate that the proposed LGSCV achieves over 20\% RD gain compared with state-of-the-art methods, while reducing 2D map generation time to approximately 1 second and cutting encoding time by 50\%. The code is available at \url{https://github.com/Qi-Yangsjtu/LGSCV}.
\end{abstract}

\section{Introduction}
3D Gaussian Splatting (GS) is the state-of-the-art (SOTA) 3D scene reconstruction technology \cite{vanillaGS}, which shows an impressive trade-off between reconstruction quality and generation speed. However, the 3D GS data, which  comprises massive explicit primitives with 59-channel features, poses a substantial challenge to effective compression. 

Most GS compression methods introduce extra constraints during the generation process, resulting in a compact representation. These methods are classified as optimization-based methods, such as HAC \cite{chen2024hac} and CompGS \cite{liu2024compgs}. They report impressive compression ratios, as well as comparable quality with vanilla GS. However, optimization-based methods tend to have long optimization and coding times on the GPU, which limits the application of 3D GS on lightweight devices \cite{m72430}. To solve the problem of lightweight GS compression, the Moving Picture Experts Group (MPEG) established the I-3DGS tracks and studied traditional GS compression methods, in which the generated 3D GS data are expected to be compressed using canonical point cloud or video codecs, such as GPCC v1 \cite{GPCCv1} and HEVC \cite{sullivan2012overview}. Considering the widespread adoption of video codecs, this work explores the use of these codecs for the efficient compression of 3D GS.

Current video-based 3D GS compression methods, such as the GSCodec Studio \cite{li2025gscodec} which is used as the anchor in the I-3DGS track, first use Parallel Linear Assignment Sorting (PLAS) \cite{PLAS} to generate smooth 2D maps based on 3D GS data, followed by feeding these maps into video codecs. PLAS is a progressive and heuristic process: it first randomly distributes the GS attributes into 2D grids, then a series of permutations are conducted to optimize the pixel distribution, making them as smooth as possible.  However, as PLAS essentially functions like a brute-force search for image smoothing, it inevitably exhibits high complexity. Using the MPEG dataset ``bartender'' which has around half a million primitives as an example, PLAS requires more than 30 seconds of GPU time.  Therefore, unless the PLAS maps are generated in advance, introducing vanilla PLAS into the coding process does not meet the requirement of being lightweight. Additionally, current video-based methods compress multidimensional GS color spherical harmonics (SH) separately at the image level, which overlooks the redundancy of high-order alternating current (AC) components \cite{yanghybridgs} and yields suboptimal performance.

In this paper, we propose LGSCV, a \textbf{L}ightweight 3D GS \textbf{C}ompression method based on \textbf{V}ideo codec, to solve the above problems.  Different from directly using PLAS to generate smooth 2D maps, we first propose a two-stage Morton scan to generate blockwise 2D maps: a 3D Morton scan is used to permute the primitives based on spatial coordinates, followed by a 2D Morton scan to map the sorted primitives to the 2D grid in a blockwise style \cite{Morton}. With the same video codec, the rate-distortion (RD) results of compressing the blockwise map are close to those of the PLAS map in the high-bitrate range, while the proposed two-stage Morton scan only requires approximately 0.2 seconds of CPU time. However, using blockwise maps demonstrates quality collapse in the medium-to-low bitstream range. Therefore, we apply principal component analysis (PCA) to reduce the SH AC feature channels and aggregate feature energy, resulting in a low-rank AC representation. By exploiting the contextual correlations among the 45-channel SH AC components, we achieve a significant RD improvement, particularly in the medium bitrate range. Finally, we propose a MiniPLAS to fine-tune the blockwise maps. By controlling the element permutation within a small block size that is similar to the coding unit (CU) of video codecs, MiniPLAS can improve compression performance in the medium-to-low bitstream range at a very fast speed: using the block size of $4*4$ with only one iteration of PLAS-style optimization, which requires around 0.8 seconds, we obtain a 2.62\% RD gain on the MPEG ``bartender'' sample. The block size of MiniPLAS can also be used as a reference to set the codec CU size. By limiting a small CU partition range, RD optimization (RDO) requires less decision time and thus significantly reduces encoding time.  The computational complexity of MiniPLAS is estimated by $O(6M^2)$ where $M^2$ is the resolution of 2D maps.  

Using the common test GS dataset suggested by MPEG, our method outperforms the SOTA video-based GS compression anchor, i.e., GSCodec Studio, by more than 20\% RD gains. We further conduct a serial ablation study to examine the potential of our method in practice. For example, the proposed LGSCV could receive an additional 3.17\% RD gain when choosing a better maximum block size (MBS) for MiniPLAS, e.g., MBS = 8, and reduce encoding time by 50\% with a smaller CU size and partition depth.  

\section{Related Work}
This section briefly reviews the related GS compression work, including optimization-based and traditional methods.

\textbf{Optimization-based Method.}
Optimization-based methods modify the generation process of vanilla GS, expecting to produce a more compact representation while maintaining comparable rendering quality. Using latent representation and grid structure, Scaffold-GS \cite{lu2024scaffold} proposed an anchor-based GS compact representation, in which each actual primitive is predicted from the anchor feature. Based on Scaffold-GS, HAC \cite{chen2024hac} was designed by introducing effective context models to estimate the probability distribution of latent features. RD loss was used as supervision to guide context model training. Another prevalent method is using codebooks and vector quantization to constrain the feature space, such as CompGS \cite{navaneet2025compgs} and C3DGS \cite{niedermayr2024compressed}. Considering the densification process can introduce redundant primitives, LightGaussian \cite{fan2023lightgaussian} and EAGLES \cite{girish2025eagles} developed primitive pruning algorithms, in which close reconstruction results are achieved by removing 60\% of the primitives followed by a fine-tuning process. To inherit the benefits of point cloud compression, HybridGS \cite{yanghybridgs} introduced quantization and sparsification during the GS generation process to obtain an explicit and compact sample. GPCC is further used to compress this sample with obviously faster encoding and decoding speeds. SOTA optimization-based methods can achieve over a 100x compression ratio with almost no quality loss \cite{xing20253dgs}. However, these methods generally require a lengthy optimization process and coding time. The coupling of generation and compression constrains the range of applications for GS.

\textbf{Traditional Method.}
The traditional method focuses on how to effectively compress the generated 3D GS data, while the optimization process based on rendering loss is eliminated from the encoding part. Inspired by the data format of 3D GS being the same as that of point clouds, MPEG WG7 first improved the GPCC v1 interface for GS data compression \cite{GPCCv1}. Based on GPCC, an adaptive voxelization is proposed for 3D GS transform coding \cite{wang2025adaptive}. Given the favorable performance of graph signal processing (GSP) on point cloud compression \cite{shaovcip} and quality assessment \cite{graphsim}, GGSC \cite{yang2024benchmark} developed a two-branch method: using GPCC to compress GS coordinates, while employing the graph Fourier transform (GFT) to decompose the GS attributes and clip the high-frequency components. HGSC \cite{huang2024hierarchical} designed a hierarchical strategy to compress the GS primitive layer by layer. FCGS \cite{chenfast} proposed the first learning-based 3D GS data compression framework via spatial grid context models. Owing to the advent of the PLAS algorithm \cite{PLAS}, using video codecs to compress 3D GS has become possible. GSCodec Studio \cite{li2025gscodec} used PLAS and FFMPEG x265 to compress GS. Traditional compression methods have a faster coding process, while the compression ratios are generally lower than optimization-based methods. 

\section{Lightweight 3D GS Compression via Video Codec}
\begin{figure}[t] {
\centering
\includegraphics[width=0.9\textwidth]{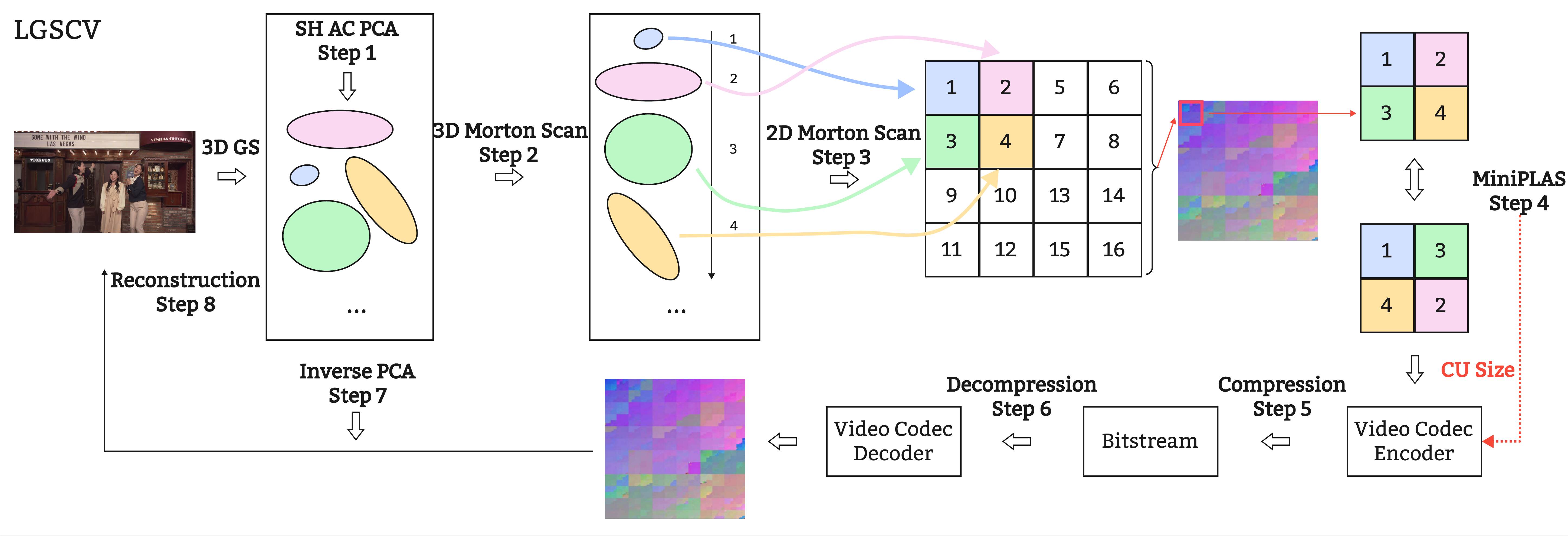}
    \caption{Framework of the proposed LGSCV.}
    \label{fig:scheme} }
    \vspace{-1.5em}
\end{figure}

Fig. \ref{fig:scheme} illustrates the overall framework of the proposed LGSCV. LGSCV consists of 8 steps, among which steps 1-5 constitute our main contribution. We will introduce them in detail in this section. 

\subsection{PCA on SH AC}
Vanilla 3D GS has 59 feature channels, of which 45 channels are color SH AC components. Current video-based methods regard these 45 channels as 15-frame RGB or 45-frame gray images, followed by video codecs. This strategy requires not only more encoding and decoding time, but also cannot fully utilize the context information between different channels, resulting in suboptimal compression results.

Inspired by optimization-based GS compression methods, SH AC components are highly redundant and thus can be reconstructed from a low-rank representation \cite{yanghybridgs}. Considering that lightweight coding is required in this paper, PCA and frequency clipping are adopted. There are three different methods to realize SH AC dimensionality reduction: 1) applying PCA once on the 45-dimensional feature matrix, then only keeping the first k components for compression while clipping the rest. 2) The 45-channel features are evenly distributed across the three color spaces, i.e., R, G, and B. We can split this matrix into three 15-channel feature matrices and use PCA separately. 3) The 15-dimensional features from each color channel belong to three orders, i.e., 3, 5, and 7. We can directly clip the higher-order without using PCA. SH AC has both intra- and inter-channel correlations; therefore, the first method is superior to the other two methods, which will be used in the proposed LGSCV.

\subsection{2D Map Generation via Two-Stage Morton Scan}
This section corresponds to steps 2 and 3 in Fig. \ref{fig:scheme}. Give spatial coordinates xyz, 3D Morton code interleaves the binary representation of the primitive, which can permute the spatially adjacent primitives into close indices \cite{Morton}. Attributes of 3D GS primitives exhibit local spatial similarity, indicating that after 3D Morton sorting, primitives with adjacent indices tend to share similar attribute values. However, using the row-by-row style to generate 2D maps is not perfectly compatible with video codec preferences, while the square block is the basic CU. Therefore, given the sorted GS primitives, we use the scan order of the 2D Morton code to generate 2D maps, as shown in the toy example in Fig. \ref{fig:scheme}. The generated 2D maps exhibit noticeable block structures, with smooth content within each block, making them well-suited for video codecs.

\subsection{MiniPLAS for Blockwise 2D Map}
This section corresponds to steps 4 and 5 of Fig. \ref{fig:scheme}. The blockwise 2D map is generated solely based on GS coordinate information, while other attributes are ignored during this process. Considering that the vanilla PLAS can use all possible attributes as a reference for sorting, we hope the PLAS can mitigate the weaknesses of the proposed two-stage Morton scan. However, directly using vanilla PLAS on the blockwise 2D map will disrupt the block structure and incur slow generation speeds, resulting in limited improvement and failing to meet the requirement of being lightweight.

Therefore, we propose a MiniPLAS that performs pixel permutation within a certain block size. We define the block sizes of permutation as $4n\times4n$, $n=1,2,4, ...$, the selected block sizes are the CU or sub-CU sizes in video codecs, such as $8\times8$.  Let $M$, $B$ denote the image resolution and block size of MiniPLAS, the complexity is $O(\lfloor \frac{M}{B} \rfloor^2\times(\frac{B^2}{4}*24))$ for a single feature channel, where $\lfloor \cdot \rfloor$ is the floor function. $\lfloor \frac{M}{B} \rfloor^2$ is the number of blocks in the 2D map, and $\frac{B^2}{4}*24$ is the number of permutation times within each block considering PLAS using 4 pixels as one unit (4! = 24 permutations for each unit). Assuming that $M\%B =0$ because $B$ is usually very small, the computational complexity can be simplified and estimated by $O(6M^2)$.

One advantage of using MiniPLAS is that it motivates the setting of the codec CU size, which can significantly reduce encoding time. For example, adaptive block partitioning is employed in HEVC, where the maximum partition depth (PD) and RDO determine the specific partition results. If $8\times8$ is used in MiniPLAS, the CU size can be located around this value, allowing the encoder to terminate further exploration of larger or smaller CUs. This approach reduces the encoding time by approximately 50\% while improving coding efficiency, as illustrated in Fig. \ref{fig:cu}.

\subsection{GS Map Compression via Video Codec}\label{sec:HM}
This section corresponds to steps 5 to 7 in Fig. \ref{fig:scheme}. For each frame of 3D GS,  we use 20 bit depth (BD) for coordinates, and 10 BD for other attributes \cite{wg7-73294}. 1, $N_k$, 1 frame images are generated for color SH DC, SH AC, and scaling, where $N_k$ depends on the PCA. For opacity and rotation, we use the mid-value of BD to pad two channels to form 1 and 2 images. To compress coordinate maps, we split the 20 BD data into two 10 BD maps. In all, $7+N_k$ 3-channel images are generated and fed into video codes to generate bitstreams. Considering that GS quality is more sensitive to coordinate information, we use lossless mode to compress coordinate images, via lossy compression for other attributes. Note that we directly regard the image data as YUV444 rather than RGB. The reasons are: 1) except for color DC and AC, there is no inter-relationship analogous to the other attributes, and 2) color space conversion between RGB and YUV is lossy, which can incur an obvious quality decrease derived from the coordinate map calculation error.

\section{Experiment Results}
We use the first frame of the three MPEG tracked sequences \cite{dataset} as the testing examples: ``bartender'', ``cinema'', and ``breakfast''. HEVC reference software (HM) version 18.0 \cite{HM} is used as the representation of the video codec. The implementation of LGSCV is in Pytorch, and the MiniPLAS is based on Vanilla PLAS \cite{PLAS-code}.  We compare the proposed LGSCV with the GSCodec Studio, which is the anchor of the MPEG GS compression working group. As suggested by \cite{N0676}, views 9 and 11 for “bartender” and “cinema”, views 6 and 8 for “breakfast” are selected to compute the RGB PSNR as quality metrics.  The results are tested on an AMD EPYC 7H12 CPU and a single NVIDIA L40s GPU.

\subsection{Overall Performance}
\begin{figure}[h]
\centering
\begin{subfigure}{0.32\textwidth}
    \includegraphics[width=\linewidth]{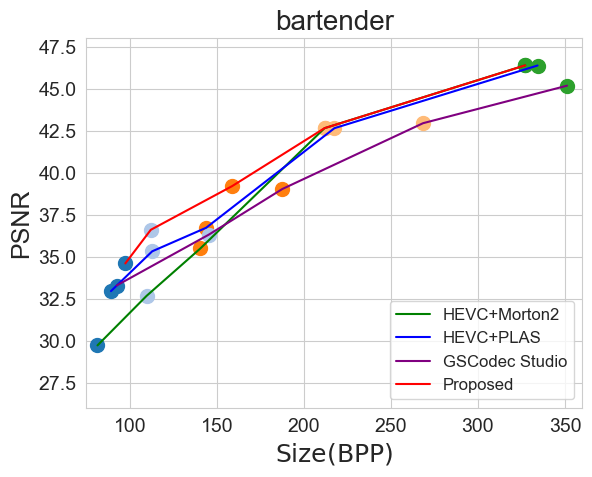}
\end{subfigure}
\begin{subfigure}{0.32\textwidth}
    \includegraphics[width=\linewidth]{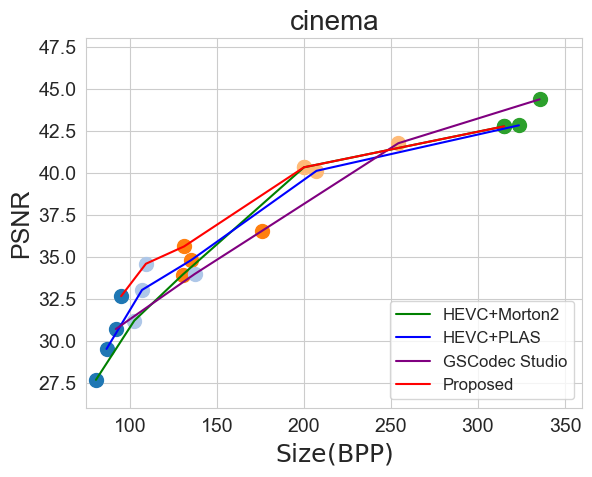}
\end{subfigure}
\begin{subfigure}{0.32\textwidth}
    \includegraphics[width=\linewidth]{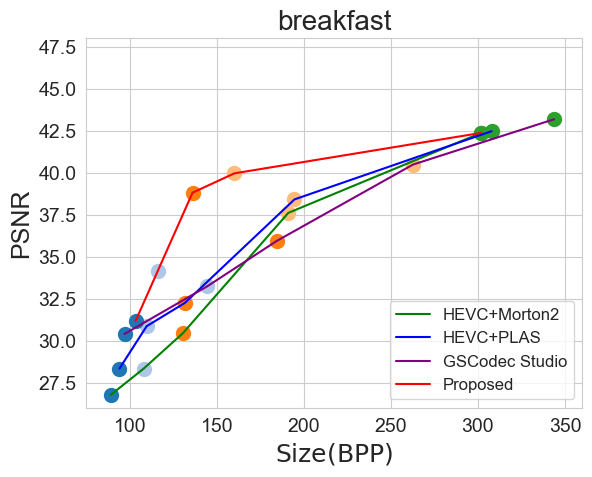}
\end{subfigure}
\caption{Performance comparison on MPEG dataset.}
\label{fig:res_overall}
\vspace{-1 em}
\end{figure}

The RD curves of the proposed LGSCV and GSCodec Studio are shown in Fig. \ref{fig:res_overall}. The selection of QPs is based on \cite{wg7-73294}, and we use the default CU setting ($64\times64$) of HEVC in this section (see section \ref{sec:ab} for other CU sizes). The first 12 components (i.e., $N_k = 4$) are used after PCA, and the miniPLAS only uses a block size of $4*4$ for one iteration of permutation. To highlight the effectiveness of the proposed modules, we also show the results of using HEVC on two different types of GS 2D maps: 1) HEVC+Morton2: using the two-stage Morton scan without PCA and MiniPLAS. 2) HEVC+PLAS: using all the GS attributes as references to perform vanilla PLAS. PCA will limit the quality upper bound, and HEVC+Morton2 is a partial module of LGSCV with disabled SH AC PCA and MiniPLAS. To demonstrate a wider bitrate range, the two highest bitrate points of ``bartender'' and ``cinema'' are the same as HEVC+Morton2, as well as the highest bitrate point of ``breakfast''. 

\begin{table}[h]
\centering
\resizebox{0.6\textwidth}{!}{
\begin{tabular}{|c|cccc|}
\hline
          & \multicolumn{4}{c|}{GSCodec   Studio}                                                            \\ \hline
Method       & \multicolumn{1}{c|}{bartender} & \multicolumn{1}{c|}{cinema} & \multicolumn{1}{c|}{breakfast} & Average \\ \hline
HEVC+Morton2 & \multicolumn{1}{c|}{-10.98}    & \multicolumn{1}{c|}{-8.75}  & \multicolumn{1}{c|}{-2.63}     & -7.45   \\ \hline
HEVC+PLAS & \multicolumn{1}{c|}{-12.55} & \multicolumn{1}{c|}{-7.21}  & \multicolumn{1}{c|}{-6.96}  & -8.91  \\ \hline
LGSCV     & \multicolumn{1}{c|}{-19.54} & \multicolumn{1}{c|}{-13.39} & \multicolumn{1}{c|}{-29.49} & -20.81 \\ \hline
\end{tabular}
}
\caption{RD gains against the GSCodec Studio using RGB PSNR measurements.}
\label{tab:RD-gain}
\vspace{-1.3em}
\end{table}

\begin{figure}[h]
\centering
\begin{minipage}{0.33\textwidth}
    \begin{subfigure}{\linewidth}
        \includegraphics[width=\linewidth]{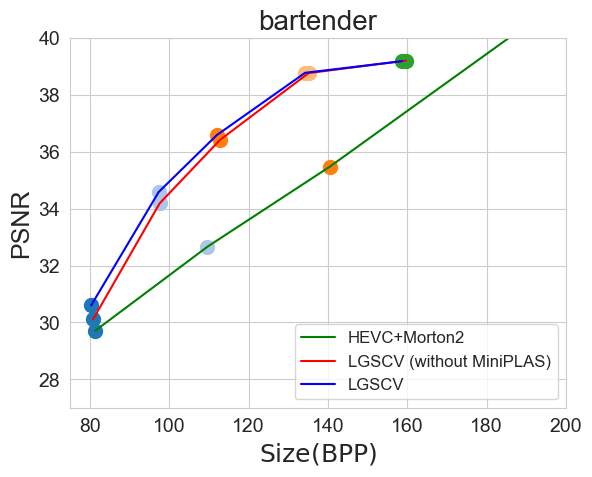}
        \caption{Ablation Study on PCA and MiniPLAS}
    \end{subfigure}\\[0.5em]
    \begin{subfigure}{\linewidth}
        \includegraphics[width=\linewidth]{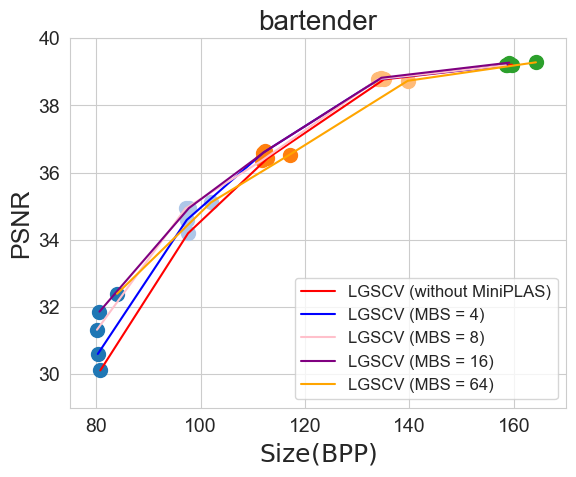}
        \caption{Ablation Study on MiniPLAS Size }
    \end{subfigure}
\end{minipage}
\hfill
\begin{minipage}{0.59\textwidth}
    \begin{subfigure}{\linewidth}
        \includegraphics[width=\linewidth]{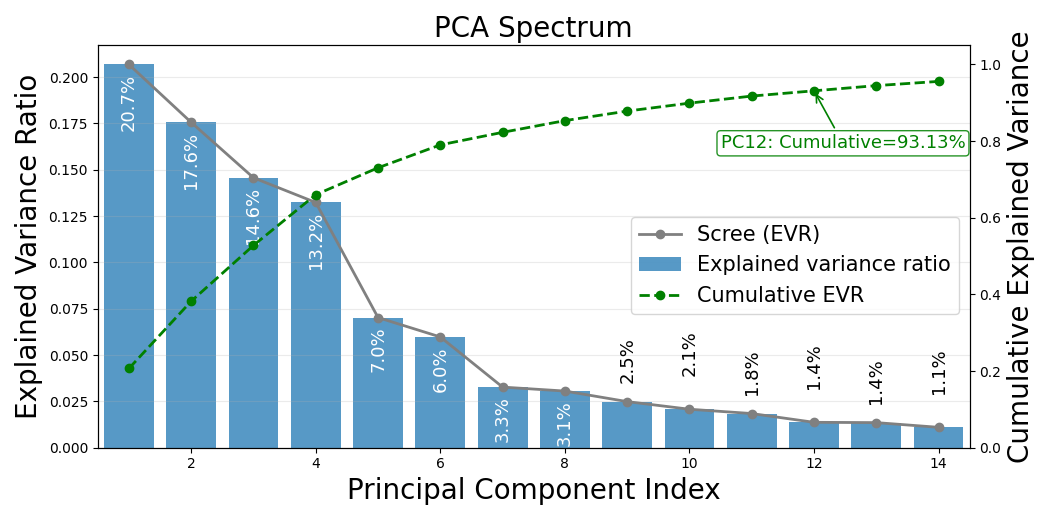}
        \caption{Bartender PCA}
    \end{subfigure}\\[0.5em]
    \begin{subfigure}{\linewidth}
        \includegraphics[width=\linewidth]{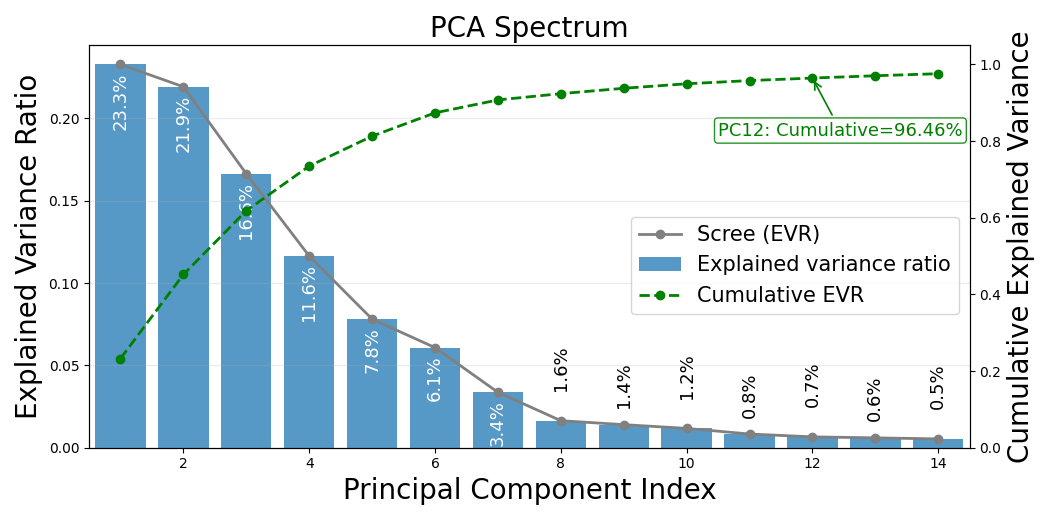}
        \caption{Breakfast PCA}
    \end{subfigure}
\end{minipage}
\caption{Ablation study of LGSCV.}
\label{fig:ab}
\vspace{-1em}
\end{figure}

We see that: 1) for the high bits per primitive (BPP) part, using HEVC+Morton2 is close to using HEVC+PLAS. Considering that the proposed two-stage Morton scan requires less than 0.1 CPU time while vanilla PLAS requires 15-35 seconds of GPU time (see Table \ref{tab:time}), it demonstrates that the proposed 2D map generation method is effective and lightweight. 2) For the medium-to-low BPP, HEVC+Morton2 is generally worse than HEVC+PLAS and GSCodec Studio. After introducing PCA and MiniPLAS, the proposed LGSCV reports improved performance over SOTA results, indicating the effectiveness of the proposed modules. The gain is mainly from using PCA as shown in Fig. \ref{fig:ab} (a). The gain from introducing PCA on ``breakfast'' is higher than that of the other two samples. The reason is that the energy ratio of the first 12 main components of ``breakfast'' is larger than that of others, as the explained variance ratio shown in Fig. \ref{fig:ab} (c) and (d): 96.46\% for ``breakfast'' while 93.13\% for ``bartender''. PCA can reduce more than 50\% of encoding and decoding time due to less input data for the video codec, and a $ 45\times12$ matrix is required as metadata for the decoding processes with PCA. 3) Table \ref{tab:RD-gain} illustrates the RD gains against the GSCodec Studio. HEVC+PLAS reports average gains of  8.91\%,  indicating that the proposed video compression anchor introduced in Section \ref{sec:HM} is more effective. The proposed LGSCV shows a 20.81\% gain, quantifying the superiority of the proposed modules. 



The generation times for the 2D maps of LGSCV and vanilla PLAS are shown in Table \ref{tab:time}. Specifically, for LGSCV, before the real coding, there are three steps: two-stage Morton scan, PCA on SH AC components, and MiniPLAS. The primitive numbers of ``bartender'', ``cinema'', and ``breakfast'' are 576,724, 423,249, and 528,154, respectively.  The proposed 3D-GS-to-2D mapping achieves a remarkably lightweight design compared with the vanilla PLAS baseline.

\begin{table}[h]
\centering
\resizebox{0.8\textwidth}{!}{
\begin{tabular}{|c|ccccc|c|}
\hline
        \multirow{2}{*}{Time(s)}  & \multicolumn{5}{c|}{LGSCV}                                                                                                & \multirow{2}{*}{PLAS} \\ \cline{2-6}
 & \multicolumn{1}{c|}{Morton 3D} & \multicolumn{1}{c|}{Morton 2D} & \multicolumn{1}{c|}{PCA} & \multicolumn{1}{c|}{MiniPLAS (MBS = 4)} & All &  \\ \hline
bartender & \multicolumn{1}{c|}{0.070} & \multicolumn{1}{c|}{0.005} & \multicolumn{1}{c|}{0.490} & \multicolumn{1}{c|}{0.810} & 1.375 & 34.400                \\ \hline
cinema    & \multicolumn{1}{c|}{0.050} & \multicolumn{1}{c|}{0.005} & \multicolumn{1}{c|}{0.580} & \multicolumn{1}{c|}{0.720} & 1.355 & 16.970                \\ \hline
breakfast & \multicolumn{1}{c|}{0.060} & \multicolumn{1}{c|}{0.005} & \multicolumn{1}{c|}{0.510} & \multicolumn{1}{c|}{0.750} & 1.325 & 24.370                \\ \hline
\end{tabular}
}
\caption{2D map generation time (seconds) of LGSCV and PLAS}
\label{tab:time}
\vspace{-2 em}
\end{table}
\subsection{Ablation Study}\label{sec:ab}
$\bullet$\textbf{MiniPLAS Block Size.} The MiniPLAS is a blockwise optimization for 2D GS maps, and the MBS determines the reception field of the element permutation. We test different MBS on LGSCV, including $64\times64$, $16\times16$, and $8\times8$. Specifically, the minimum block size is defined as $4\times4$. For example, when MBS is 64, it means we perform five iterations progressively with block sizes [64, 32, 16, 8, 4], where we reduce the block size to a quarter after each iteration. The results are shown in Fig. \ref{fig:ab} (b). We see that against the no MiniPLAS cases,  the performance is improved for MBS=4 to MBS=16, with gains of 2.62\%, 6.05\%, and 3.37\%, while it is reduced when MBS=64, with -0.92\%. The improvements are primarily observed in the medium and low bitrate regions. The reason is that using MiniPLAS strikes a balance between the coordinates and other attributes in the bitstreams. After the two-stage Morton scan, the coordinate map is very smooth, while other attributes are relatively rough. Introducing the MiniPLAS will damage the smoothness of coordinate maps, while other attribute maps are improved. Considering that coordinate maps are lossless compression while others are lossy, using a too large  MBS will significantly increase the coordinate bitstream. One potential improvement is to design a better loss function \cite{MPED} to guide pixel permutation.

\begin{wrapfigure}{r}{0.40\textwidth} 
  \vspace{-10pt} 
  \centering
  \includegraphics[width=0.40\textwidth]{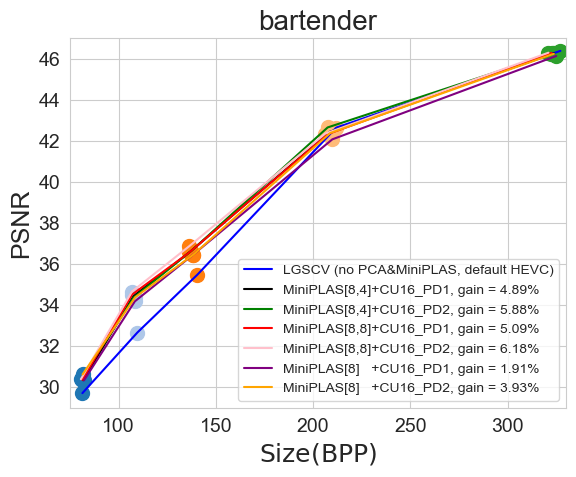}
  \caption{Ablation Study on CU}
  \label{fig:cu}
  \vspace{-12pt} 
\end{wrapfigure}

\noindent$\bullet$\textbf{CU Size.} Fig. \ref{fig:cu} exhibits the results of using different MiniPLAS and HEVC CU/PD settings. We disable PCA for generality. [8, 4] denotes using two iterations with block sizes 8 and 4 in MiniPLAS, CU16 means the CU size is $16\times16$, and PD1 means the maximum PD is 1; the others follow the same pattern. The default HEVC is CU64 and PD4. The performance differences of PD1 and PD2 are around 1\%-2\% corresponding to different MiniPLAS settings. The average per-frame encoding times of default, CU16\_PD2, and CU16\_PD1 are 5.9, 4.5, and 2.27 seconds. MiniPLAS with $8\times8$ requires approximately 1 second, indicating that it can also be regarded as a powerful tool that benefits both the compression ratio and encoding efficiency.

\section{Conclusion}
In this paper, we propose LGSCV, a lightweight 3D GS compression method based on video codecs. There are three main contributions in the proposed LGSCV. First, considering PLAS is time-consuming for 2D GS map generation, we propose a two-stage Morton scan to generate blockwise 2D maps, which is lightweight and shows comparable performance to using PLAS in the high bitstream range. Second, to better utilize contextual information between high-dimensional color SH AC components, we use PCA to generate a low-rank representation for SH AC. It improves RD results in the medium bitstream part, as well as reduces the data volume required for encoding. Third, following the blockwise style of the 2D map, a MiniPLAS is developed to realize fast and effective primitive permutation within certain block sizes. With only a few iterations, MiniPLAS improves the RD results in the low bitstream and increases coding speed by guiding CU size selection. The proposed LGSCV reports a 20.81\% average gain over the MPEG anchor on the common test dataset and requires less than 1.5 seconds to generate the desired 2D maps.

\section{References}
\bibliographystyle{IEEEtran}
\bibliography{refs}

\end{document}